\documentclass[10pt,twocolumn,letterpaper]{article}

\usepackage{cvpr}              
\usepackage{algorithm}
\usepackage{algorithmicx}
\usepackage{algpseudocode}
\usepackage{amsmath}
\usepackage{amssymb}
\usepackage{graphicx}

\usepackage{hyperref}
\usepackage{multirow}
\usepackage{dirtree}
\usepackage{caption}  
\usepackage[accsupp]{axessibility}  

\definecolor{cvprblue}{rgb}{0.21,0.49,0.74}

\def\paperID{39} 
\def\confName{CVPR}
\def\confYear{2026}

\title{FruitEnsemble: MLLM-Guided Arbitration for Heterogeneous ensemble in Fine-Grained Fruit Recognition}

\author{
Enhui Yu\textsuperscript{1}, 
Junhui Li\textsuperscript{1,2}, 
Ruitong Lu\textsuperscript{1}, 
Jialu Li\textsuperscript{3}, 
Youshan Zhang\textsuperscript{2,*} \\
\textsuperscript{1}\textit{University of Science and Technology Liaoning}, 
\textsuperscript{2}\textit{Chuzhou University},
\textsuperscript{3}\textit{Yeshiva University}\\
*Corresponding author, youshan\_zhang@chzu.edu.cn
}
\begin{document}
\maketitle
\begin{abstract}

Fine-grained fruit classification is a critical yet challenging task in agricultural computer vision, primarily hindered by a severe shortage of high-quality datasets and the high visual similarity between classes. To address these challenges, we first constructed a comprehensive dataset comprising 306 fruit categories with 116,233 samples. Moreover, we propose FruitEnsemble, a practical two-stage dynamic inference framework designed to overcome the generalization limitations of static single-model architectures. In the first stage, FruitEnsemble employs a validation-calibrated weighted ensemble of heterogeneous backbones to generate a robust Top-3 candidate pool. To tackle difficult samples, we introduce an expert arbitration mechanism: when ensemble confidence falls below 0.6, a multimodal large language model (MLLM) is triggered to perform rigorous visual verification by integrating external botanical descriptions using Chain-of-Thought (CoT) reasoning. Furthermore, we optimized the training pipeline with a hard sample-aware joint loss. Extensive experiments demonstrate that FruitEnsemble achieves a classification accuracy of 70.49\% and outperforms existing state-of-the-art models. Our framework provides an efficient, deployment-oriented solution for real-world agricultural visual sorting and quality inspection tasks.\footnote{This research was funded by the Research Projects of Anhui Province Education Department (Grant No. 2025AHGXZK20009).} 
\end{abstract}
    
\section{Introduction}
\label{sec:intro}

Accurate fruit cultivar recognition plays a crucial role in modern agricultural supply chains, directly influencing quality assurance, pricing, and post-harvest storage~\cite{garg2023smart}. 
Beyond these core tasks, reliable identification also supports downstream applications like automated sorting, yield estimation, and inventory management.

In practice, fruit classification has long relied on manual inspection and expert judgment. While workable at a small scale, manual sorting is time-consuming, labor-intensive, costly, and often inconsistent, making it difficult to meet the increasing demand for automation in large-scale agricultural systems. Image-based fruit classification has emerged as an attractive alternative because it is non-destructive, and easily integrated into intelligent sorting lines, harvesting robots, and agricultural monitoring platforms.

\begin{figure}[tbp]
  \centering
  \includegraphics[width=\columnwidth]{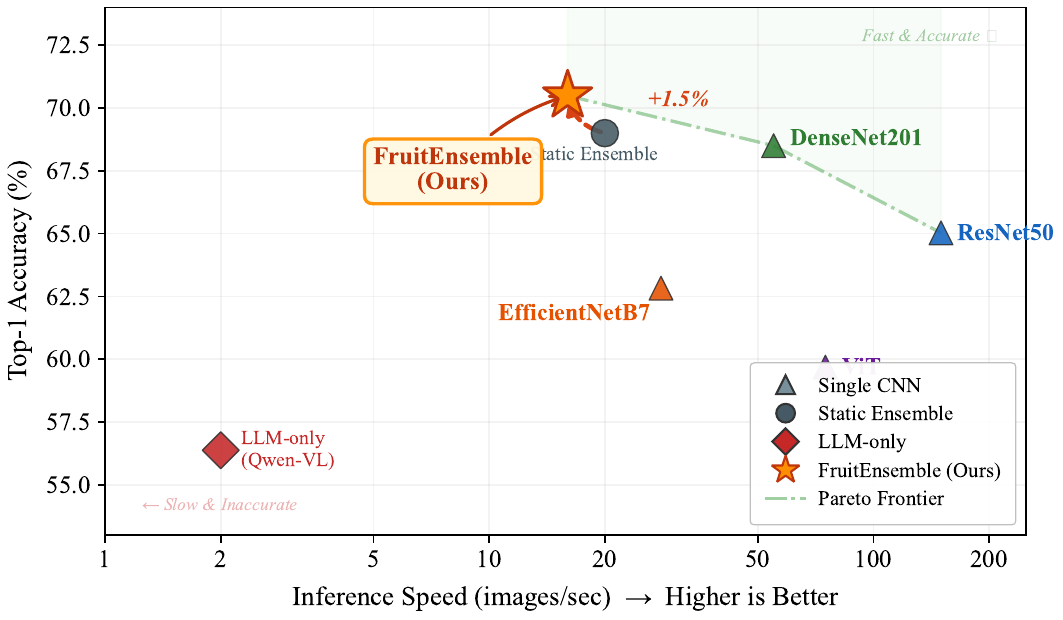}
  \caption{\textbf{Accuracy vs.\ inference efficiency} on the Fruit-306 test set. 
  Single CNNs (\textcolor[HTML]{78909C}{$\boldsymbol{\triangle}$}) are efficient but limited in accuracy. 
  The LLM-only baseline (\textcolor[HTML]{C62828}{$\boldsymbol{\diamond}$}) is slow and performs poorly without domain-specific training. 
  Static ensembles (\textcolor[HTML]{455A64}{$\bullet$}) improve accuracy but remain sub-optimal. 
  FruitEnsemble (\textcolor[HTML]{FF8F00}{$\bigstar$}) achieves the best trade-off by invoking the LLM only for uncertain samples.}
  \label{fig:method_comparison}
  \vspace{-0.cm}
\end{figure}

Despite recent advances in modern visual backbones, fruit cultivar recognition remains a challenging fine-grained visual classification problem~\cite{meng2021deep}. Cultivars such as Red Fuji and Gala may appear very similar in color, shape, and size, yet they must be reliably distinguished in applications such as automated sorting and quality control. The challenge arises from both subtle inter-class differences and significant intra-class variation caused by factors such as ripeness, illumination, viewpoint, and occlusion. Additionally, agricultural datasets are often long-tailed, with a few common cultivars dominating the training data while many rare ones are underrepresented~\cite{du2025improving}. Existing visual methods are efficient and relatively mature, but a single model is often not robust enough across all fine-grained categories. Ensemble methods can improve robustness~\cite{dietterich2000ensemble}, yet they still process every sample in the same way and cannot adapt their computation to sample difficulty.

Vision-language models have recently shown strong capabilities in image understanding and cross-modal reasoning. However, in fruit classification, prior work has focused primarily on pure visual recognition based on CNNs~\cite{krizhevsky2012imagenet}, Vision Transformers~\cite{dosovitskiy2020image}, or their ensembles. The use of language-guided reasoning with category-level descriptions for fine-grained fruit recognition remains relatively underexplored. As a result, neither pure visual classification nor indiscriminate use of large models is ideal: the former can struggle with fine-grained and long-tailed cases, while the latter is expensive to deploy and requires sufficient domain grounding for reliable predictions~\cite{yao2024survey}. As shown in Fig.~\ref{fig:method_comparison}, existing paradigms still face a clear trade-off between recognition accuracy and inference efficiency, which motivates our approach.

To address these challenges, we propose \textbf{FruitEnsemble}, a dynamic two-stage framework for fine-grained fruit classification. In the first stage, four complementary visual models -- ResNet50~\cite{he2016deep}, DenseNet201~\cite{huang2017densely}, EfficientNetB7~\cite{tan2019efficientnet}, and Vision Transformer -- are combined through confidence-aware weighted fusion to produce a prediction and a confidence score. In the second stage, a multimodal LLM (Qwen-VL-Plus) is invoked only when the ensemble is uncertain and performs reasoning over the top candidate classes with the aid of curated category descriptions~\cite{zhang2025illusionbench}. This design allows the easiest samples to be handled efficiently by the visual ensemble, while reserving more expensive language-guided reasoning for genuinely ambiguous cases. Our contributions are three-fold:

\begin{itemize}

    \item We construct \textbf{Fruit-306}, a large-scale fine-grained fruit recognition benchmark with 306 categories and 116,233 images. Uniquely, it pairs visual data with expert-curated \textit{textual morphological descriptions}, enabling rigorous evaluation of knowledge-guided multimodal reasoning in agriculture.
    
    \item We present \textbf{FruitEnsemble}, a practical two-stage recognition framework that combines validation-calibrated heterogeneous ensembling with selective MLLM-based arbitration. By constraining the MLLM to a \textit{Top-$K$} candidate set and providing category-specific botanical descriptions, the system improves hard-case verification while avoiding unrestricted open-set generation.
 
    \item We propose a \textit{Hard Sample-Aware Joint Optimization} strategy. By applying a specific \textbf{diversity loss} only to hard samples that the router has detected, we force backbones to learn complementary features where needed, while also optimizing the training process. This enables SOTA \textbf{70.49\% accuracy} with real-time latency (\textbf{19.8 ms}), invoking the LLM for only 15\% of cases.
\end{itemize}

\section{Related Work}
\label{sec:related_work}

\subsection{Fine-Grained Fruit Recognition}

Fine-grained visual recognition has been studied extensively in domains such as birds and cars. Early approaches emphasized part discovery and alignment~\cite{zhang2014part}, later methods relied on attention mechanisms~\cite{fu2017look}, and recent transformer-based architectures have shown strong performance~\cite{he2022transfg}. However, transferring these methods to fruit variety recognition poses unique challenges: cultivar-level differences are rarely expressed through stable semantic parts but instead through weak cues such as texture, gloss, stripe density, and color transition~\cite{hou2017dualnet}, all of which are easily affected by lighting, occlusion, and growth stage. Deep learning has been applied to apple cultivar classification~\cite{bhargava2021classification} and citrus grading~\cite{xu2025research}, yet most work addresses coarse-grained separation among a small number of varieties and lacks sufficient discriminative power when facing hundreds of visually similar cultivars.

Multimodal large language models such as Qwen-VL-Plus and GPT-4V~\cite{yang2023dawn} provide a unified interface for visual understanding and language-based reasoning, and can generate interpretable decisions through chain-of-thought prompting~\cite{wei2022chain}. Initial studies have evaluated the zero-shot capability of MLLMs for crop disease diagnosis~\cite{wang2025large}, but their performance on cultivar-level fine-grained discrimination remains unstable, and inference latency makes them impractical as standalone high-throughput classifiers.
\subsection{Agricultural Vision Benchmarks}

The benchmark landscape further compounds this limitation. PlantVillage~\cite{noyan2022uncovering} focuses on disease recognition, and Fruits-360~\cite{oltean2017fruits} provides images under controlled backgrounds for fruit-level classification; neither captures the difficulty of variety-level recognition under realistic conditions. In particular, existing public benchmarks rarely combine a large number of similar cultivar categories, substantial intra-class variation — the latter being essential for vision-language joint reasoning and a gap that our dataset is specifically designed to address.

\subsection{Ensemble Model}

Ensemble learning improves robustness by combining predictions from multiple models~\cite{li2012diversity} and has been applied to crop disease recognition~\cite{astani2022diverse} and fruit quality grading. However, conventional ensembles are static: every sample is processed by every model with fixed fusion rules. In high-throughput scenarios such as automated fruit sorting lines, this uniform computation is inefficient since most samples are easy to classify. Adaptive inference methods such as early-exit networks~\cite{teerapittayanon2016branchynet} and dynamic ensemble selection~\cite{cruz2015meta} adjust computation per sample, but they are mostly limited to homogeneous model families and have not been systematically applied to agricultural fine-grained recognition, where lightweight classifiers and stronger reasoning models need to collaborate.

Our work targets closed-set recognition over hundreds of visually similar fruit varieties and combines three elements: a heterogeneous ensemble of complementary vision backbones, a confidence-based dynamic escalation mechanism, and category-level textual descriptions for semantic grounding during MLLM reasoning. The MLLM serves as a selective arbiter for ambiguous samples rather than a replacement for specialized visual classifiers.
\section{Fruit-306 Dataset}
\label{sec:dataset}

We create \textbf{Fruit-306}, a benchmark for fine-grained fruit recognition in realistic settings. The dataset contains \textbf{116,233} images from \textbf{306} categories and is designed to support both standard visual classification and vision-language reasoning. Compared with existing fruit datasets, Fruit-306 is characterized by three aspects that are central to this work: a larger set of fine-grained categories, real-world visual complexity, and category-level textual descriptions. Images are collected under diverse conditions and include cluttered backgrounds, viewpoint changes, occlusion, and different ripeness stages, making the benchmark substantially more challenging than other datasets.

Fruit-306 adopts a fixed \textbf{7:1:2} split into training, validation, and test sets, containing \textbf{81,233}, \textbf{11,488}, and \textbf{23,522} images, respectively, while preserving the original class distribution. As shown in Table~\ref{tab:dataset_stats}, the dataset exhibits a clear long-tail distribution: the largest category contains \textbf{1,276} images (\textit{jaboticaba}), while the smallest contains only \textbf{25} images (\textit{muscadine\_grape}), resulting in an imbalance ratio of \textbf{50:1}. This naturally preserved long-tail distribution reflects real-world agricultural data availability and requires reliable recognition for both frequent and rare categories.

\begin{table}[h]
    \centering
    \caption{Statistical breakdown of the \textbf{Fruit-306} dataset. The partition maintains the original long-tail distribution across training, validation, and testing sets. Note the significant variance in class sizes, ranging from abundant commercial varieties to rare species.}
    \label{tab:dataset_stats}
    \resizebox{\columnwidth}{!}{%
    \begin{tabular}{lcccccc}
        \toprule
        \textbf{Split} & \textbf{Ratio} & \textbf{\# Images} & \textbf{\# Classes} & \textbf{Max Sample Class} & \textbf{Min Sample Class} & \textbf{Avg. Size} \\
        \midrule
        Training & 70\% & 81,223 & \multirow{3}{*}{306} & \multirow{3}{*}{\shortstack{1,276 \\ (\textit{jaboticaba})}} & \multirow{3}{*}{\shortstack{25 \\ (\textit{muscadine\_grape})}} & \multirow{3}{*}{379.85} \\
        Validation & 10\% & 11,488 & & & & \\
        Testing & 20\% & 23,522 & & & & \\
        \midrule
        \textbf{Total} & \textbf{100\%} & \textbf{116,233} & \textbf{306} & \multicolumn{3}{c}{\textit{Imbalance Ratio (Max/Min): 50:1}} \\
        \bottomrule
    \end{tabular}%
    }
\end{table}

Beyond image labels, each category is paired with a short expert-curated textual description that summarizes its discriminative visual traits, including morphology, color patterns, texture, and local structure. These descriptions provide category-level priors that are useful when visual evidence alone is insufficient. In our framework, they serve as semantic guidance for the language model when the visual ensemble is uncertain, directly supporting the dynamic inference strategy introduced in Sec.~\ref{sec:method}.

\begin{figure*}[t]
  \centering

  \includegraphics[width=\textwidth]{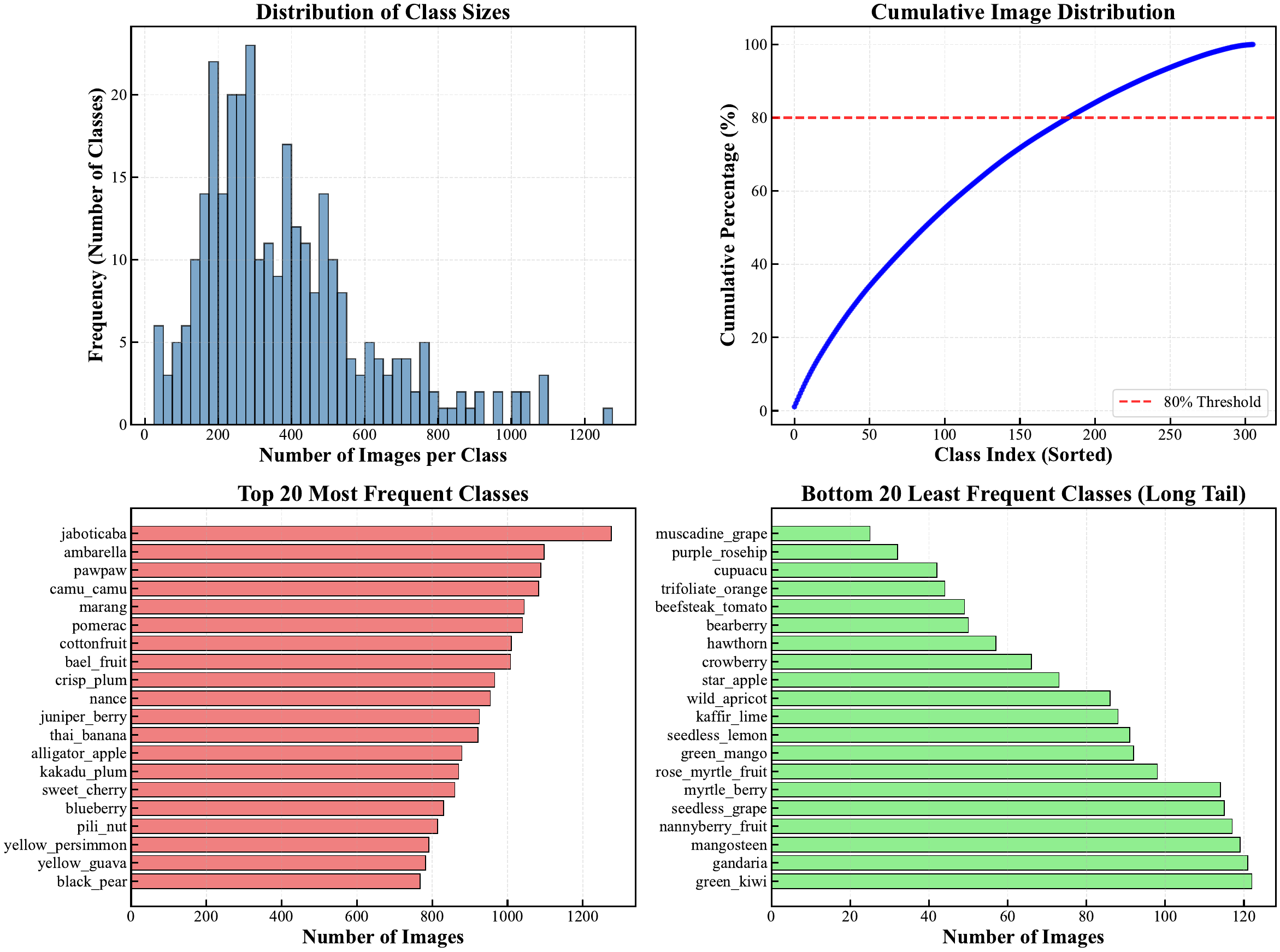}
  
  \vspace{-0.3cm}
  \caption{\textbf{Comprehensive Statistical Analysis of Fruit-306.} 
  \textbf{(a)} Histogram of class sizes showing the frequency distribution of samples per category. 
  \textbf{(b)} Cumulative distribution curve indicating that the top 20\% of classes account for a significant majority of the total images. 
  \textbf{(c)} Top 20 most frequent classes (Head), dominated by common commercial varieties. 
  \textbf{(d)} Bottom 20 least frequent classes (Tail), highlighting the data scarcity in rare species. 
  These statistics confirm the severe long-tail nature of the dataset.}
  \label{fig:detailed_stats}
\end{figure*}

The class distribution is summarized in Fig.~\ref{fig:detailed_stats}. The histogram and cumulative curve confirm a severe long-tail pattern, with a small number of head classes accounting for a large fraction of the data. This imbalance, together with real-world appearance variation, makes Fruit-306 a more demanding benchmark than balanced fruit datasets.

The dataset follows standard PyTorch \texttt{ImageFolder} format, with folders for training, validation, and test splits. This structure is used in all experiments.


Table~\ref{tab:dataset_comparison} compares Fruit-306 with representative fruit and VegFru~\cite{hou2017vegfru} datasets. Fruit-306 combines fine-grained category coverage, realistic image conditions, natural long-tail imbalance, and textual descriptions in a single benchmark. This combination is particularly suitable for evaluating the proposed method, which integrates visual experts with language-guided reasoning.

\begin{table}[ht]
  \centering
  \caption{\textbf{Comparison of Fruit-306 with existing fruit related datasets.} Our dataset leads in classes, realism, and multimodal annotations (Desc.: Description).}
  \label{tab:dataset_comparison}
  \resizebox{\linewidth}{!}{
  \begin{tabular}{lccccc}
    \toprule
    Dataset & \# Classes & \# Images & Real-World & Long-Tail & Text Desc. \\
    \midrule
    Fruits-360~\cite{kaggle_fruits} & 131 & 90k & No & No & No \\
    VegFru~\cite{hou2017vegfru} & 292 & 160k+ & Yes & Yes & No \\
    FruitVision~\cite{bijoy2025fruitvision} & 5 & 10,154 & Yes & No & No \\
    \textbf{Fruit-306 (Ours)} & \textbf{306} & \textbf{116,233}& \textbf{Yes} & \textbf{Yes} & \textbf{Yes} \\
    \bottomrule
  \end{tabular}
  }
\end{table}

\begin{figure}[t]
  \centering

  \includegraphics[width=0.95\linewidth]{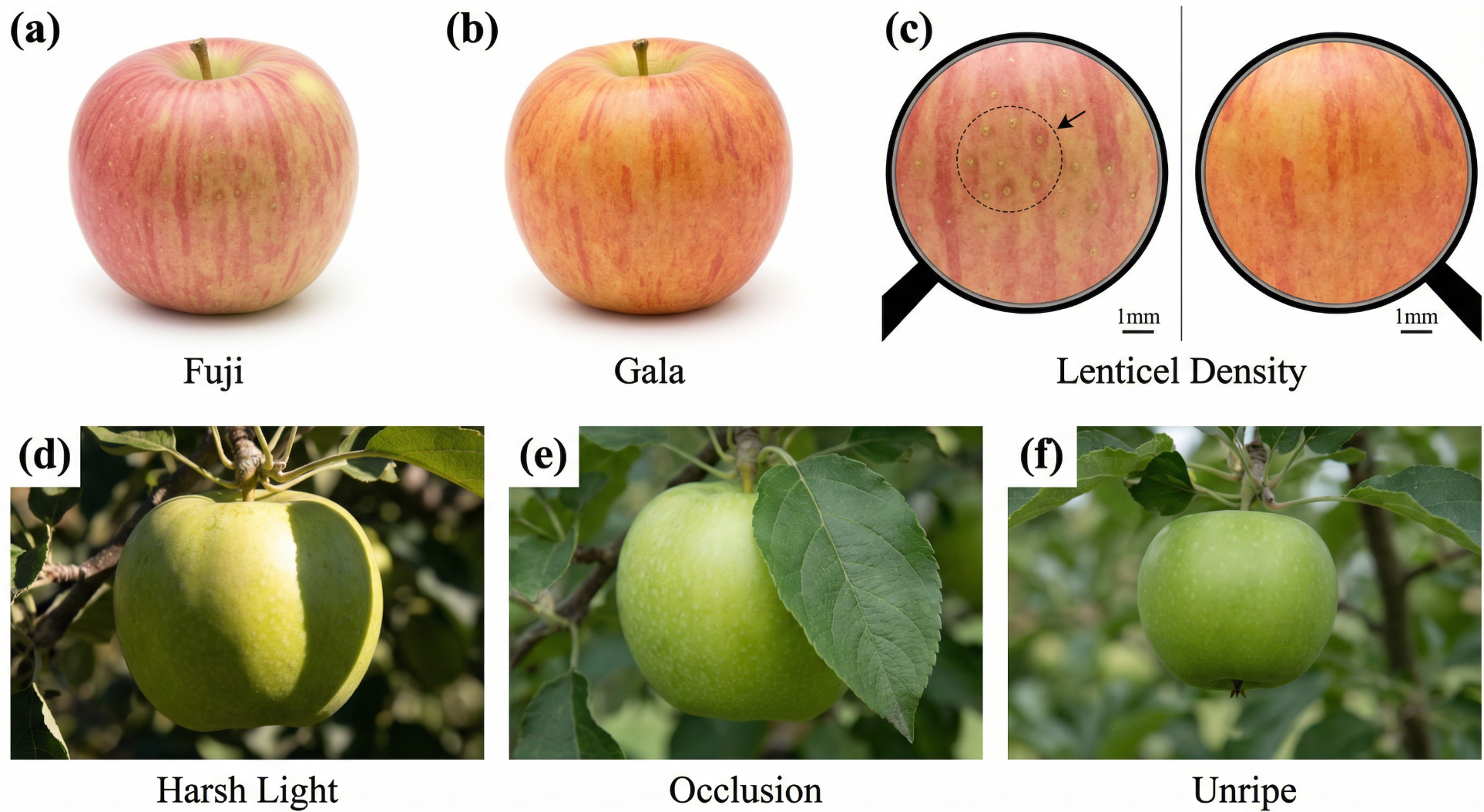}

  \vspace{-0.3cm} 
  
  \caption{\textbf{Visual Challenges in Fruit-306.} 
  \textbf{Top Row:} High \textit{inter-class similarity}. Examples of visually similar pairs (e.g., \textit{Fuji} vs. \textit{Gala}) where discrimination relies on subtle textual attributes (e.g., lenticel density). 
  \textbf{Bottom Row:} High \textit{intra-class variance}. The same variety shown under different lighting, occlusion, and ripeness conditions. These complexities necessitate dynamic reasoning beyond static classification.}
  
  \label{fig:dataset_samples}
\end{figure}

From the recognition perspective, Fruit-306 is challenging for four reasons: high inter-class similarity, large intra-class variation, cluttered natural backgrounds, and severe class imbalance. Figure~\ref{fig:dataset_samples} shows typical examples. These factors motivate the design of \textbf{FruitEnsemble}: easy samples are efficiently classified by the ensemble vision model, while ambiguous cases benefit from language-guided reasoning with category descriptions. We evaluate this strategy on the fixed split of Fruit-306 in Sec.~\ref{sec:experiments}.

\section{Method}
\label{sec:method}

\begin{figure*}[t]
    \centering

    \includegraphics[width=\textwidth]{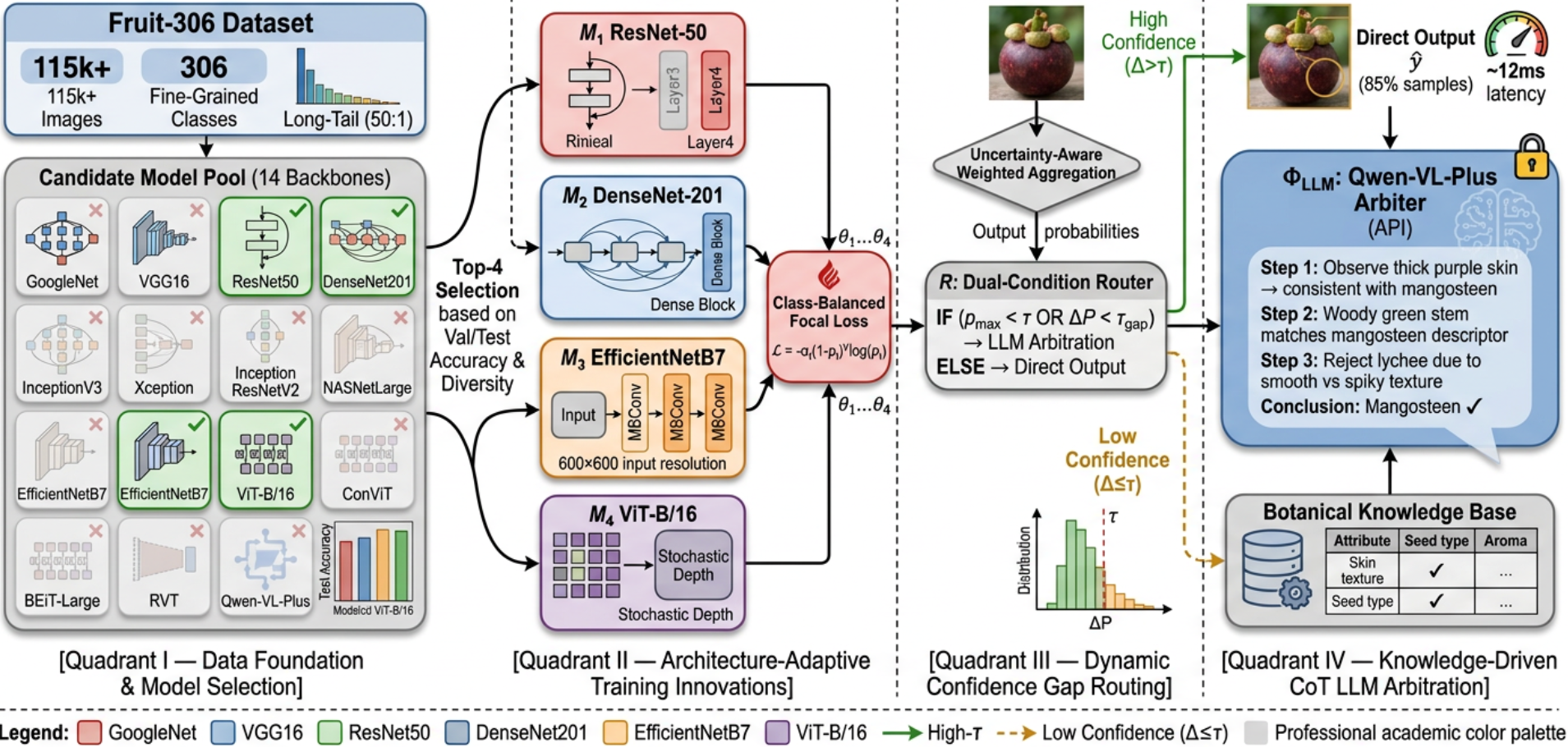}
    \caption{\textbf{Overview of the FruitEnsemble Framework.} 
    Input fruit images are first processed by a curated set of four heterogeneous backbone models ($\mathcal{M}_1$-$\mathcal{M}_4$: ResNet50, DenseNet201, EfficientNetB7, ViT-B/16) with architecture-adaptive training strategies. 
    Their output probabilities are fused via an \textit{Uncertainty-Aware Weighted Aggregation} module ($\mathcal{A}$) to generate top-$k$ candidate predictions and a confidence gap metric ($\Delta = p_1 - p_2$). 
    The \textit{Confidence-Gap Router} ($\mathcal{R}$) applies a dual-condition rule: if the maximum confidence ($p_{\text{max}} > \tau$) \textit{and} confidence gap ($\Delta > \tau_{\text{gap}}$), samples follow the direct path for immediate prediction ($\hat{y}$); ambiguous samples (low confidence or small margin) are routed to the \textit{Qwen-VL-Plus LLM Arbiter} ($\Phi_{\text{LLM}}$). 
    The LLM performs Knowledge-Augmented Chain-of-Thought (CoT) reasoning using curated fruit descriptor knowledge to produce the final error-corrected prediction ($\hat{y}^*$).}
    \label{fig:framework}
\end{figure*}

Fine-grained fruit variety recognition faces two intertwined challenges: visually similar cultivars demand strong discriminative power, while long-tail class distributions and high-throughput deployment requirements impose strict efficiency constraints. To address this trade-off, we present \textbf{FruitEnsemble}, a hierarchical collaborative framework that uses complementary inductive biases of heterogeneous visual architectures for robust feature learning. We only apply computationally heavy LLM reasoning to ambiguous examples that cannot be resolved by visual information alone.
Formally, let $\mathcal{X}$ denote the input image space and $\mathcal{Y} = \{1, \dots, C\}$ denote the set of $C=306$ fruit categories. 
Our framework $\mathcal{F}$ is defined as a tuple $(\mathcal{M}, \mathcal{A}, \mathcal{R}, \Phi_{\text{LLM}})$, comprising: 
(1) a heterogeneous backbone ensemble $\mathcal{M}$; 
(2) an uncertainty-aware aggregation operator $\mathcal{A}$; 
(3) a \textit{confidence-gap triggered} router $\mathcal{R}$;
(4) a Top-$K$ constrained LLM arbiter $\Phi_{\text{LLM}}$.
The overall pipeline is illustrated in \textbf{Figure~\ref{fig:framework}}.

\subsection{Uncertainty-Aware Heterogeneous Fusion}
\label{subsec:ensemble}

To mitigate the inductive bias inherent in single architectures, we construct an ensemble $\mathcal{M} = \{m_1, m_2, m_3, m_4\}$ consisting of four distinct backbones~\cite{dietterich2000ensemble}, selected based on their empirical complementarity (see Table~\ref{tab:backbone_stats}):
\begin{itemize}
    \item \textbf{DenseNet201}~\cite{huang2017densely} ($m_1$): excels in capturing fine-grained texture patterns (e.g., skin speckles) due to dense connectivity.
    \item \textbf{EfficientNetB7}~\cite{tan2019efficientnet} ($m_2$): provides superior multi-scale representation, yielding high Top-5 recall.
    \item \textbf{Vision Transformer}~\cite{dosovitskiy2020image} ($m_3$): models long-range dependencies and global morphology, offering an orthogonal perspective to CNNs.
    \item \textbf{ResNet50}~\cite{he2016deep} ($m_4$): serves as a robust baseline with balanced hierarchical feature extraction.
\end{itemize}

\noindent\textbf{Architecture-Specific Adaptation.} 
To ensure optimal convergence of these heterogeneous components, we employ tailored training strategies for each backbone, including \textit{topology-aware layer freezing}~\cite{he2016deep}, \textit{gradient clipping with non-finite handling~\cite{graves2013generating}}, and \textit{activation-matched classifier heads}. These techniques stabilize training and maximize the individual capacity of each model, forming a robust foundation for the ensemble. Detailed implementation protocols are provided in Sec.~\ref{sec:experiments}.

Given an input image $x$, each backbone $m_i$ outputs a class probability distribution $P_i(x)$. 
Instead of static averaging, we propose an \textit{Uncertainty-Aware Dynamic Aggregation} mechanism. 
Let $\mathcal{U}_i(x) = -\sum_{c} P_i(x)_c \log P_i(x)_c$ denote the predictive entropy~\cite{gal2016dropout}. 
The dynamic weight $\alpha_i(x)$ is computed as:
\begin{equation}
    \alpha_i(x) = \frac{\exp\left(-\mathcal{U}_i(x) / \tau_{temp}\right)}{\sum_{j=1}^{4} \exp\left(-\mathcal{U}_j(x) / \tau_{temp}\right)},
    \label{eq:dynamic_weights}
\end{equation}
where $\tau_{temp}$ is a temperature parameter~\cite{hinton2015distilling}. The aggregated system distribution is $P_{\text{sys}}(x) = \sum_{i=1}^{4} \alpha_i(x) P_i(x)$.
From $P_{\text{sys}}(x)$, we extract the Top-$K$ ($K=3$) candidate set $\mathcal{C}_{\text{top}}(x) = \{c_1, c_2, c_3\}$, where $c_1$ is the most probable class.

\subsection{Dynamic Confidence-Gap Arbitration Routing}
\label{subsec:router}

A distinguishing feature of FruitEnsemble is the \textit{Dynamic Confidence-Gap Arbitration}. 
Relying solely on the maximum probability $S(x) = P_{\text{sys}}(x)_{c_1}$ can be misleading: a model might be confidently wrong, or it might be highly uncertain between two visually similar classes (e.g., two apple varieties) despite having a high $P_{c_1}$. 
To capture this ambiguity, we define the \textbf{Confidence Gap}~\cite{bartlett2008classification} $\Delta(x)$ as the margin between the top two predictions:
\begin{equation}
    S(x) = P_{\text{sys}}(x)_{c_1}, \quad \Delta(x) = P_{\text{sys}}(x)_{c_1} - P_{\text{sys}}(x)_{c_2}.
    \label{eq:metrics}
\end{equation}
The routing function $\mathcal{R}(x)$ triggers the LLM arbitration path if the sample is either \textit{uncertain} (low $S(x)$) or \textit{ambiguous} (small $\Delta(x)$):
\begin{equation}
    \mathcal{R}(x) = \mathbb{I}\left( S(x) < \tau_{conf} \lor \Delta(x) < \tau_{gap} \right),
    \label{eq:router_def}
\end{equation}
where $\tau_{conf}$ and $\tau_{gap}$ are empirically determined thresholds. 
This dual-criterion ensures both low-confidence and \textit{highly contested} samples are escalated for deeper reasoning, preventing the ensemble from making hasty decisions on fine-grained boundaries~\cite{fu2017look}.

The final prediction $\hat{y}$ is determined by:
\begin{equation}
    \hat{y} = 
    \begin{cases} 
      c_1, & \text{if } \mathcal{R}(x) = 0 \quad (\text{Direct Path}) \\
      \Phi_{\text{LLM}}(x, \mathcal{C}_{\text{top}}(x)), & \text{if } \mathcal{R}(x) = 1 \quad (\text{Arbitration Path}).
    \end{cases}
    \label{eq:final_prediction}
\end{equation}
Empirically, this strategy routes approximately 15\% of the most challenging samples to the LLM, optimizing the cost-accuracy trade-off.

\subsection{Top-$K$ Constrained Chain-of-Thought Arbitration}
\label{subsec:llm_arbitration}

For samples routed to the arbitration path, the LLM acts not as a generative classifier over the entire label space $\mathcal{Y}$, but as a verifier within the restricted hypothesis space $\mathcal{C}_{\text{top}}(x)$. 
This constraint reduces hallucination risks by limiting the solution space to the most probable candidates identified by the visual ensemble.
Specifically, let $\mathcal{C}_{\text{top}}(x) = \{c_1, \dots, c_K\}$ denote the top-$K$ candidate set output by the preceding stage, where $K \ll |\mathcal{Y}|$.
We construct the LLM prompt using the input image, the candidate set $\mathcal{C}_{\text{top}}(x)$, and botanical priors $\mathcal{T}=\{T_c\}_{c\in \mathcal{C}_{\text{top}}}$, where each $T_c$ is an expert-curated textual description of a candidate fruit category.
The LLM $\Phi_{\text{LLM}}$ performs candidate-aware reasoning:
\begin{enumerate}
    \item \textbf{Visual Attribute Extraction}: Identify discriminative features $A = \text{Extract}(V(x))$ (e.g., skin glossiness, calyx structure)~\cite{radford2021learning}.
    \item \textbf{Prior Matching \& Exclusion}: Compute semantic alignment scores $M_c = \text{Similarity}(A, T_c)$ and explicitly penalize contradictions:
    \begin{equation}
        \hat{y}_{\text{llm}} = \operatorname*{arg\,max}_{c \in \mathcal{C}_{\text{top}}} \left( M_c - \lambda_{\text{pen}} \cdot \text{Conflict}(A, T_c) \right).
        \label{eq:llm_deduction}
    \end{equation}
\end{enumerate}

\textbf{Implementation Details}: To operationalize this reasoning process, our system dynamically retrieves the corresponding expert botanical priors from our curated text database. Rather than relying on a static prompt, these retrieved descriptions are injected into a standardized Chain-of-Thought (CoT) template. The MLLM is explicitly instructed to execute the aforementioned extraction and matching steps, finally outputting the decision in a strict JSON format to prevent parsing errors and hallucinations. For the complete prompt templates, extensive database snippets, and full reasoning case studies, please refer to our project repository: \url{https://mybkgjvgnd.github.io/Fruit-306-Dataset/}.

\subsection{Hard Sample-Aware Joint Optimization}
\label{subsec:loss}

Training such a heterogeneous system requires balancing individual accuracy with collective diversity, particularly for long-tail classes. 
We propose a \textit{Hard Sample-Aware Joint Loss}~\cite{lin2017focal} $\mathcal{L}_{\text{total}}$:
\begin{equation}
    \mathcal{L}_{\text{total}} = \underbrace{\sum_{i=1}^4 \mathcal{L}_{\text{focal}}(P_i, y)}_{\mathcal{L}_{\text{ind}}} + \lambda_1 \underbrace{\mathcal{L}_{\text{focal}}(P_{\text{sys}}, y)}_{\mathcal{L}_{\text{global}}} + \lambda_2 \underbrace{\mathcal{L}_{\text{div}}^{\text{hard}}}_{\mathcal{L}_{\text{div}}},
    \label{eq:total_loss}
\end{equation}
where $\mathcal{L}_{\text{focal}}$ addresses class imbalance.

The critical component is the diversity term $\mathcal{L}_{\text{div}}^{\text{hard}}$, which applies Jensen-Shannon divergence maximization \textit{only} to samples identified as hard by the router ($\mathcal{R}(x)=1$):
\begin{equation}
    \mathcal{L}_{\text{div}}^{\text{hard}} = \frac{1}{|\mathcal{B}_{\text{hard}}|} \sum_{x \in \mathcal{B}_{\text{hard}}} \left[ -\sum_{i \neq j} \text{JS}(P_i(x) \parallel P_j(x)) \right],
    \label{eq:ohem_loss}
\end{equation}
where $\mathcal{B}_{\text{hard}} = \{x \in \mathcal{B} \mid \mathcal{R}(x) = 1\}$. 
\textbf{Rationale:} Applying diversity loss to easy samples may destabilize the consensus on clear patterns. By restricting it to hard samples, we specifically force the ensemble to explore complementary feature subspaces (e.g., texture vs. shape) for ambiguous cases. This ensures that the ground truth label is highly likely to be included in the Top-$K$ candidate set $\mathcal{C}_{\text{top}}(x)$, thereby maximizing the success rate of the subsequent LLM arbitration.

\subsection{Robust Training Algorithm}
\label{subsec:training_algo}

We propose a \textit{Robust High-Resolution Fine-Tuning} strategy (Algorithm~\ref{alg:unified_robust_training}) that integrates heterogeneous adaptation with fault-tolerant optimization. 
Key features include: 
(1) \textbf{Topology-Aware Freezing}~\cite{yosinski2014transferable}, which unfreezes only final decision blocks to preserve generic features; 
(2) \textbf{Fault-Tolerant Flow}, employing NaN-aware batch skipping and non-finite gradient clipping to prevent collapse on noisy data; 
(3) \textbf{EMA Regularization}, maintaining weight averages as a temporal ensemble;~\cite{zhang2019lookahead} 
(4) \textbf{Dynamic Scheduling} via Cosine Annealing with Warm Restarts to escape local minima.

\begin{algorithm}[t]
\footnotesize 
\caption{Robust High-Resolution Fine-Tuning}
\label{alg:unified_robust_training}
\begin{algorithmic}[1]
\Require Dataset $\mathcal{D}$, Backbones $\{\Theta_{base}^{(i)}\}$, Resolutions $\{R_i\}$
\Ensure Optimized Ensemble $\{\Theta^{*(i)}\}$, EMA Shadows $\{\Theta_{ema}^{(i)}\}$
\State \textbf{Phase 1: Adaptation}
\For{each backbone $i$ \textbf{do}}
    \State Resize inputs to $R_i$; Unfreeze final block \& classifier
    \State Init $\Theta^{(i)} \leftarrow \Theta_{base}^{(i)}$, $\Theta_{ema}^{(i)} \leftarrow \Theta^{(i)}$
\EndFor
\State \textbf{Phase 2: Optimization} ($\mathcal{L}_{total}$, AdamW, Cosine Scheduler)
\While{epoch $e < E_{max}$ \textbf{do}}
    \For{batch $(X, y)$ \textbf{do}}
        \State Compute outputs $\{P_i\}$, $P_{sys}$; Identify hard samples $\mathcal{B}_{hard}$
        \State $\mathcal{L} \leftarrow \mathcal{L}_{total}(\{P_i\}, P_{sys}, y, \mathcal{B}_{hard})$
        \If{$\mathcal{L}$ is NaN/Inf} \State Skip batch; \textbf{continue} \EndIf
        \State $\mathcal{L}$.backward() (with accumulation)
        \If{update step \textbf{then}}
            \State Clip gradients (tolerant); Optimizer step(); Zero grad
            \For{each $i$ \textbf{do}} \State Update EMA: $\Theta_{ema}^{(i)} \leftarrow \lambda \Theta_{ema}^{(i)} + (1-\lambda)\Theta^{(i)}$ \EndFor
        \EndIf
    \EndFor
\EndWhile
\State \Return Best checkpoints
\end{algorithmic}
\end{algorithm}

\subsection{Computational Efficiency}
\label{subsec:efficiency}
Concerns regarding LLM-augmented ensemble latency are mitigated by our \textit{adaptive computation}. 
Let $T_{CNN}$ and $T_{LLM}$ denote inference times for backbones and the LLM, respectively. 
Unlike naive ensembles ($4T_{CNN} + T_{LLM}$), our Dynamic Confidence-Gating invokes the LLM only for a fraction $\gamma \approx 0.15$ of samples. 
The expected latency is $E[T] = 4T_{CNN} + \gamma T_{LLM}$. 
Since backbones run in parallel and $\gamma \ll 1$, FruitEnsemble achieves near-ensemble accuracy with latency comparable to a single model, ensuring viability for real-time deployment.
\section{Experiments}
\label{sec:experiments}

\subsection{Experimental Setup}
\label{sec:exp_setup}

All experiments were conducted on a workstation equipped with an NVIDIA GeForce RTX 5060 Ti GPU using PyTorch with CUDA acceleration. 
To construct a heterogeneous ensemble with complementary representation capabilities, four widely used backbones were selected: 
ResNet50, DenseNet201, EfficientNetB7, and Vision Transformer (ViT-B/16). 
These architectures represent diverse design paradigms in fine-grained visual classification (FGVC): 
ResNet50 provides robust hierarchical feature extraction, and DenseNet201 captures fine-grained texture patterns through dense connectivity. 
EfficientNetB7 offers a strong multi-scale representation with compound scaling, 
and Vision Transformer models long-range dependencies via global self-attention.

Training was performed for 100 epochs using the AdamW optimizer with an initial learning rate of $5\times10^{-5}$ and weight decay of 0.01. 
The batch size was set to 8. 
To address class imbalance in the dataset, Focal Loss ($\gamma=2.0$) was employed, where the class balancing parameter $\alpha$ was set inversely proportional to class sample frequency. 
Input images were resized to $224\times224$ for most models, while adhering to the original architecture specifications, $600\times600$ for EfficientNetB7, followed by normalization to the range [0,1]. 
Standard data augmentation techniques were applied, including random horizontal flipping, random cropping, color jittering, and $\pm15^\circ$ rotation.

Different fine-tuning strategies were applied to each backbone to better adapt them to the target task. 
For ResNet50, the final residual block (layer4) and classification head were unfrozen for training. 
DenseNet201 unfroze the final dense block and applied an exponential moving average (EMA) to stabilize training. 
EfficientNetB7 fine-tuned the last MBConv block with cosine annealing scheduling, 
while Vision Transformer employed stochastic depth regularization to improve generalization.

Model performance was evaluated using Top-1 accuracy, Top-5 accuracy, and macro-F1 score on validation and test sets. 
Additionally, performance across three fruit quality grades (high-quality, defective, and flawed) was analyzed. 
For the ensemble framework, the aggregation temperature was set to 1.0, the routing confidence threshold was 0.60, and the candidate set size was $K=3$. 
The LLM arbitrator (Qwen-VL-Plus) was prompted using chain-of-thought (CoT) reasoning, with maximum generation length 512 and temperature 0.7. 
A response caching mechanism was used to reduce repeated inference latency.

To facilitate reproducibility and future research, the source code and the proposed Fruit-306 dataset are available at: \url{https://mybkgjvgnd.github.io/Fruit-306-Dataset/}.
\subsection{Experimental Results}
\label{sec:exp_results}

Table~\ref{tab:backbone_stats} presents a comprehensive comparison of 14 individual backbone models together with the proposed FruitEnsemble framework. 
The four heterogeneous backbones selected for the ensemble are highlighted in bold. DenseNet201 achieves the highest individual Top-1 accuracy, while EfficientNetB7 and ViT demonstrate superior Top-5 recall, highlighting their complementary characteristics. FruitEnsemble leverages heterogeneous ensemble aggregation to surpass all individual models. 

\begin{table}[t]
\centering
\caption{\textbf{Comprehensive Baseline Performance Analysis.} Comparison of 14 individual backbones and the proposed FruitEnsemble. The \textbf{four selected heterogeneous backbones} are highlighted in bold.\textsuperscript{\dag} Average API latency per request. 
\textsuperscript{\ddag} Average system latency (85\% direct path).}
\label{tab:backbone_stats}

\renewcommand{\arraystretch}{1.6} 

\setlength{\tabcolsep}{2.5pt} 

\resizebox{\linewidth}{!}{%
\begin{tabular}{lcccccc}
\toprule[\heavyrulewidth]

\textbf{Model} & \textbf{Top-1 Acc} & \textbf{Top-5 Acc} & \textbf{Params (M)} & \textbf{FLOPs (G)} & \textbf{Latency (ms)} \\
\midrule[\heavyrulewidth]

GoogleNet~\cite{szegedy2015going} & 0.6240 & 0.8164 & 6.8 & 1.5 & 8.2 \\
VGG16~\cite{simonyan2014very} & 0.5860 & 0.7595 & 138.0 & 15.5 & 22.4 \\
Inceptionv3~\cite{szegedy2016rethinking} & 0.1635 & 0.2763 & 23.8 & 5.7 & 14.1 \\
Xception~\cite{chollet2017xception} & 0.5830 & 0.7493 & 22.9 & 8.4 & 16.5 \\
InceptionResnetV2~\cite{szegedy2017inception} & 0.6734 & 0.8062 & 55.9 & 13.2 & 25.6 \\
NasnetLarge~\cite{zoph2018learning} & 0.2228 & 0.3914 & 88.9 & 23.6 & 38.2 \\
\midrule

\textbf{ResNet50}~\cite{he2016deep} & \textbf{0.6503} & \textbf{0.8658} & 25.6 & 4.1 & 12.5 \\
\textbf{DenseNet201}~\cite{huang2017densely} & \textbf{0.6850} & \textbf{0.8802} & 20.0 & 4.4 & 18.2 \\
\textbf{EfficientNetB7}~\cite{tan2019efficientnet} & \textbf{0.6283} & \textbf{0.8916} & 66.3 & 37.0 & 45.6 \\
\textbf{ViT-B/16}~\cite{dosovitskiy2020image} & \textbf{0.5969} & \textbf{0.8831} & 86.6 & 17.6 & 32.4 \\
\midrule

CONVT~\cite{wu2021cvt} & 0.5802 & 0.7836 & 28.4 & 6.2 & 15.8 \\
Beit\_large~\cite{bao2021beit} & 0.5373 & 0.6744 & 304.0 & 63.4 & 55.2 \\
RVT~\cite{messina2022recurrent} & 0.5347 & 0.7229 & 102.5 & 24.1 & 36.5 \\
\midrule

Qwen-VL-Plus~\cite{wang2024qwen2} & 0.5638 & - & API & API & 1250\textsuperscript{\dag} \\
\midrule[\heavyrulewidth]

\textbf{FruitEnsemble (Ours)} & \textbf{0.7049} & \textbf{0.9150} & \textit{--} & \textit{--} & \textbf{19.8}\textsuperscript{\ddag} \\
\bottomrule[\heavyrulewidth]
\end{tabular}
}

\vspace{0.8em}
\noindent\footnotesize

\end{table}

Among individual models, DenseNet201 achieves the highest Top-1 accuracy (0.6850), indicating its strong ability to capture fine-grained texture details. 
EfficientNetB7 and ViT-B/16 demonstrate the best Top-5 accuracy (0.8916 and 0.8831), suggesting their effectiveness in capturing broader semantic cues. 
In contrast, several traditional models, such as Inceptionv3, show significantly degraded performance due to limited adaptability to high intra-class variance.

The multimodal model Qwen-VL-Plus achieves 0.5638 Top-1 accuracy but suffers from extremely high latency (1250 ms), making it unsuitable for real-time applications. 
In comparison, the proposed FruitEnsemble achieves 0.7049 Top-1 accuracy and 0.9150 Top-5 accuracy while maintaining an average latency of only 19.8 ms. 
Overall, FruitEnsemble improves the baseline accuracy by \textbf{+5.0\% absolute} (65.5\% → 70.5\%), demonstrating the effectiveness of the proposed hybrid architecture.

\subsection{Ablation Study}
\label{sec:ablation}

\begin{table}[h]
\centering
\caption{\textbf{Ablation Study on FruitEnsemble Components.}}
\label{tab:ablation}
\resizebox{\linewidth}{!}{
\begin{tabular}{lccc|c}
\toprule
\textbf{Configuration} & \textbf{Heterogeneous} & \textbf{TTA} & \textbf{LLM Arb.} & \textbf{Top-1 Acc (\%)} \\
\midrule
Baseline (ResNet50~\cite{he2016deep}) & -- & -- & -- & 65.03 \\
+ Heterogeneous Ensemble & \checkmark & -- & -- & 68.50 \\
+ Test-Time Augmentation & \checkmark & \checkmark & -- & 68.92 \\
+ LLM Arbitration (Ours) & \checkmark & \checkmark & \checkmark & \textbf{70.49} \\
\bottomrule
\end{tabular}
}
\\
\footnotesize{All configurations use the same backbone parameters. The LLM is triggered only for samples with confidence $< 0.60$.}
\end{table}

\begin{figure}[t]
\centering
\includegraphics[width=\linewidth]{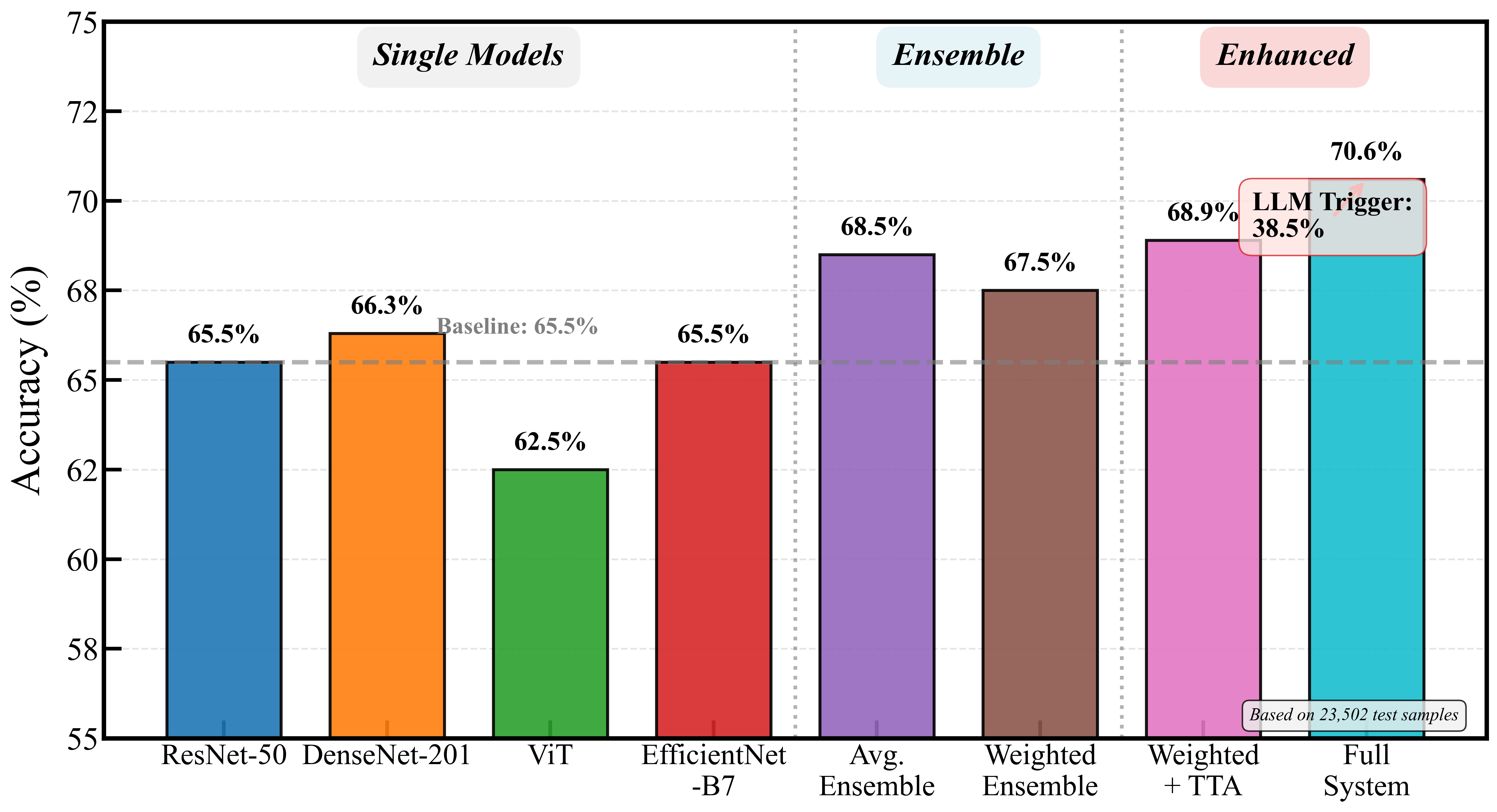}
\caption{\textbf{Progressive performance improvement of FruitEnsemble.} 
Starting from individual backbones, heterogeneous ensemble aggregation, test-time augmentation, and LLM arbitration progressively improve accuracy from 65.5\% to 70.5\%.}
\label{fig:ablation_progress}
\end{figure}

To better understand the contribution of each module, we conduct ablation experiments on the key components of FruitEnsemble. 
Figure~\ref{fig:ablation_progress} illustrates the progressive improvement path from individual backbones to the final integrated system. 
Starting from the ResNet50 baseline (65.5\% Top-1 accuracy), heterogeneous ensemble modeling significantly improves performance to 68.5\%, confirming the complementarity between CNN and transformer architectures. 
Introducing test-time augmentation further increases accuracy to 68.9\%, while the LLM-based arbitration mechanism resolves ambiguous cases and achieves the final 70.5\% accuracy.

\begin{figure}[t]
\centering
\includegraphics[width=\linewidth]{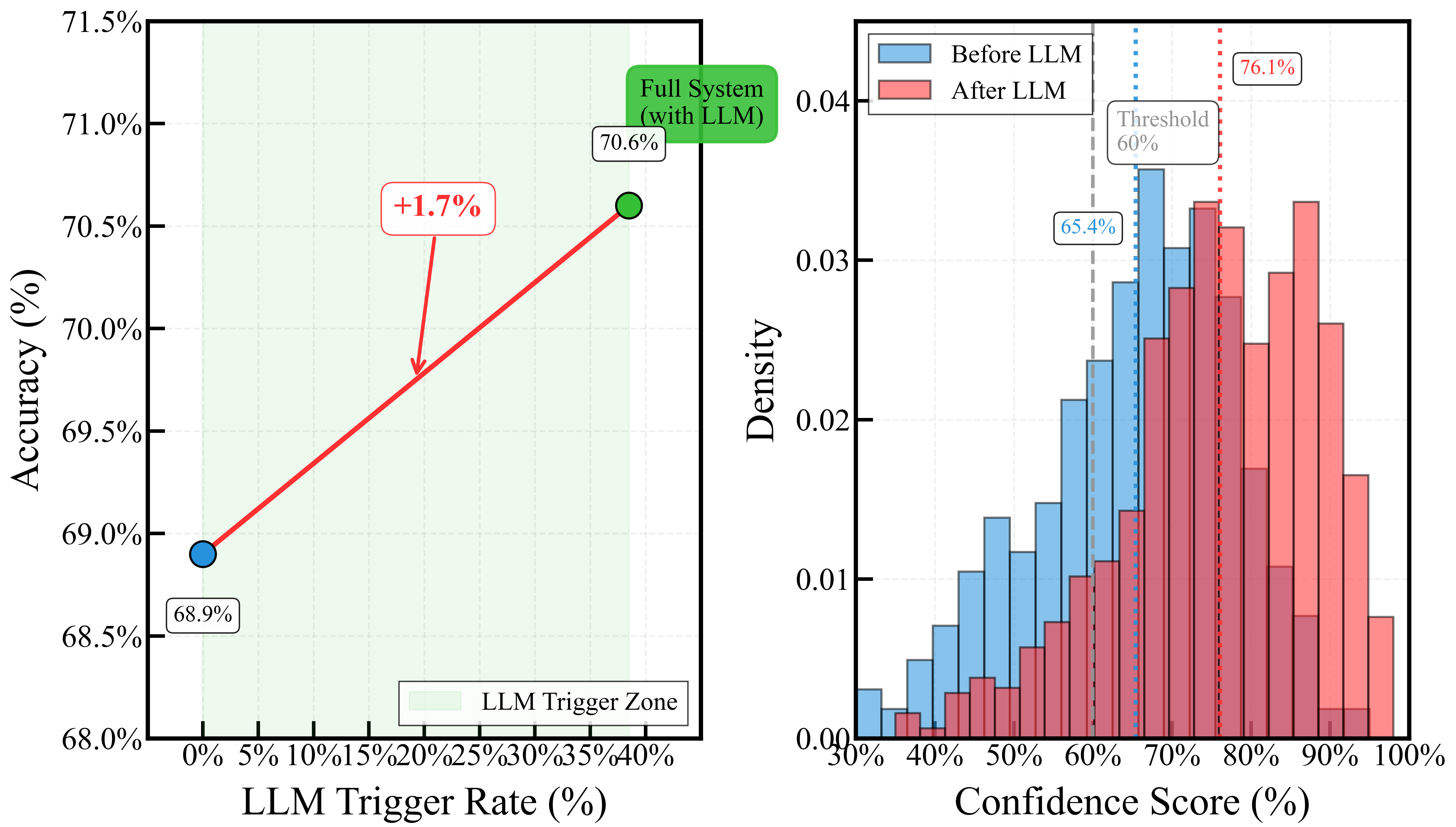}
\caption{\textbf{Impact of LLM trigger rate on classification accuracy based on test samples.} 
As the LLM trigger rate increases from 0\% to 40\%, accuracy improves marginally from 68.90\% to 70.49\%, 
indicating that LLM arbitration provides limited gains on this dataset. 
The optimal trigger rate is around 38.5\%, balancing computational cost and performance improvement.}
\label{fig:llm_trigger_analysis}
\end{figure}

Figure~\ref{fig:llm_trigger_analysis} illustrates the effect of  LLM trigger thresholds on overall accuracy. 
The relatively flat curve suggests that the base ensemble model has high confidence in most samples, 
and LLM intervention primarily benefits a small subset of difficult cases. 
This analysis helps determine the optimal threshold for deploying LLM arbitration in practice, trading off between accuracy gains and computational overhead.

\section{Conclusion}
\label{sec:conclusion}
To meet the requirements of \textbf{accurate classification and real-time efficiency} for agricultural automated sorting, this study constructed the Fruit-306 dataset containing 306 fruit categories and more than 116K images captured in real-world scenarios, and proposed the FruitEnsemble dynamic collaborative framework, which is specially designed to address the core challenges of \textbf{high cultivar similarity, large intra-class variation and imbalanced data distribution}.
Through an uncertainty-aware two-stage mechanism, visual models efficiently process easy samples while the LLM arbitrates only a small number of ambiguous cases, achieving an optimal balance between accuracy and efficiency with an average latency of merely 19.8 milliseconds.
Experimental results show that classification accuracy is significantly improved, with a Top-1 accuracy of 70.49\% that outperforms all single models and static ensembles. The framework can accurately identify visually similar fruit cultivars and exhibits markedly enhanced robustness in real-world agricultural environments with complex lighting, occlusion, and varying ripeness levels. 

{
    \small
    \bibliographystyle{ieeenat_fullname}
    \bibliography{main}

@article{meng2021deep,
  title={Deep learning for fine-grained classification of jujube fruit in the natural environment},
  author={Meng, Xi and Yuan, Yingchun and Teng, Guifa and Liu, Tianzhen},
  journal={Journal of Food Measurement and Characterization},
  volume={15},
  number={5},
  pages={4150--4165},
  year={2021},
  publisher={Springer}
}

@article{du2025improving,
  title={Improving long-tailed pest classification using diffusion model-based data augmentation},
  author={Du, Mengze and Wang, Fei and Wang, Yu and Li, Kun and Hou, Wenhui and Liu, Lu and He, Yong and Wang, Yuwei},
  journal={Computers and Electronics in Agriculture},
  volume={234},
  pages={110244},
  year={2025},
  publisher={Elsevier}
}

@inproceedings{he2016deep,
  title={Deep residual learning for image recognition},
  author={He, Kaiming and Zhang, Xiangyu and Ren, Shaoqing and Sun, Jian},
  booktitle={Proceedings of the IEEE conference on computer vision and pattern recognition},
  pages={770--778},
  year={2016}
}

@inproceedings{huang2017densely,
  title={Densely connected convolutional networks},
  author={Huang, Gao and Liu, Zhuang and Van Der Maaten, Laurens and Weinberger, Kilian Q},
  booktitle={Proceedings of the IEEE conference on computer vision and pattern recognition},
  pages={4700--4708},
  year={2017}
}

@inproceedings{tan2019efficientnet,
  title={Efficientnet: Rethinking model scaling for convolutional neural networks},
  author={Tan, Mingxing and Le, Quoc},
  booktitle={International conference on machine learning},
  pages={6105--6114},
  year={2019},
  organization={PMLR}
}

@article{dosovitskiy2020image,
  title={An image is worth 16x16 words: Transformers for image recognition at scale},
  author={Dosovitskiy, Alexey and Beyer, Lucas and Kolesnikov, Alexander and Weissenborn, Dirk and Zhai, Xiaohua and Unterthiner, Thomas and Dehghani, Mostafa and Minderer, Matthias and Heigold, Georg and Gelly, Sylvain and others},
  journal={arXiv preprint arXiv:2010.11929},
  year={2020}
}

@misc{kaggle_fruits,
  author       = {Mihai Gheorghe},
  title        = {Fruits 360: A Dataset of Images Containing Fruits and Vegetables},
  year         = {2018},
  publisher    = {Kaggle},
  howpublished = {\url{https://www.kaggle.com/datasets/moltean/fruits}},
  note         = {Accessed: 2024-05-20}
}

@inproceedings{hou2017vegfru,
  title={Vegfru: A domain-specific dataset for fine-grained visual categorization},
  author={Hou, Saihui and Feng, Yushan and Wang, Zilei},
  booktitle={Proceedings of the IEEE international conference on computer vision},
  pages={541--549},
  year={2017}
}

@article{bijoy2025fruitvision,
  title     = {FruitVision: A benchmark dataset for fresh, rotten, and formalin-mixed fruit detection},
  author    = {Bijoy, Md Hasan Imam and others},
  journal   = {Data in Brief},
  volume    = {61},
  pages     = {111752},
  year      = {2025},
  publisher = {Elsevier},
  doi       = {10.1016/j.dib.2025.111752},
  pmcid     = {PMC12221502},
  pmid      = {40605851}
}

@inproceedings{hou2017dualnet,
  title={Dualnet: Learn complementary features for image recognition},
  author={Hou, Saihui and Liu, Xu and Wang, Zilei},
  booktitle={Proceedings of the IEEE international conference on computer vision},
  pages={502--510},
  year={2017}
}

@inproceedings{dietterich2000ensemble,
  title={Ensemble methods in machine learning},
  author={Dietterich, Thomas G},
  booktitle={International workshop on multiple classifier systems},
  pages={1--15},
  year={2000},
  organization={Springer}
}

@article{yao2024survey,
  title={A survey on large language model (llm) security and privacy: The good, the bad, and the ugly},
  author={Yao, Yifan and Duan, Jinhao and Xu, Kaidi and Cai, Yuanfang and Sun, Zhibo and Zhang, Yue},
  journal={High-Confidence Computing},
  volume={4},
  number={2},
  pages={100211},
  year={2024},
  publisher={Elsevier}
}

@inproceedings{zhang2025illusionbench,
  title={Illusionbench: A large-scale and comprehensive benchmark for visual illusion understanding in vision-language models},
  author={Zhang, Yiming and Zhang, Zicheng and Wei, Xinyi and Liu, Xiaohong and Zhai, Guangtao and Min, Xiongkuo},
  booktitle={2025 IEEE International Conference on Multimedia and Expo (ICME)},
  pages={1--6},
  year={2025},
  organization={IEEE}
}

@inproceedings{zhang2014part,
  title={Part-based R-CNNs for fine-grained category detection},
  author={Zhang, Ning and Donahue, Jeff and Girshick, Ross and Darrell, Trevor},
  booktitle={European conference on computer vision},
  pages={834--849},
  year={2014},
  organization={Springer}
}

@inproceedings{fu2017look,
  title={Look closer to see better: Recurrent attention convolutional neural network for fine-grained image recognition},
  author={Fu, Jianlong and Zheng, Heliang and Mei, Tao},
  booktitle={Proceedings of the IEEE conference on computer vision and pattern recognition},
  pages={4438--4446},
  year={2017}
}

@inproceedings{he2022transfg,
  title={Transfg: A transformer architecture for fine-grained recognition},
  author={He, Ju and Chen, Jie-Neng and Liu, Shuai and Kortylewski, Adam and Yang, Cheng and Bai, Yutong and Wang, Changhu},
  booktitle={Proceedings of the AAAI conference on artificial intelligence},
  volume={36},
  number={1},
  pages={852--860},
  year={2022}
}

@article{noyan2022uncovering,
  title={Uncovering bias in the PlantVillage dataset},
  author={Noyan, Mehmet Alican},
  journal={arXiv preprint arXiv:2206.04374},
  year={2022}
}

@article{oltean2017fruits,
  title={Fruits 360 dataset on github},
  author={Oltean, Mihai and Muresan, Horea},
  year={2017}
}

@inproceedings{li2012diversity,
  title={Diversity regularized ensemble pruning},
  author={Li, Nan and Yu, Yang and Zhou, Zhi-Hua},
  booktitle={Joint European conference on machine learning and knowledge discovery in databases},
  pages={330--345},
  year={2012},
  organization={Springer}
}

@article{cruz2015meta,
  title={META-DES: A dynamic ensemble selection framework using meta-learning},
  author={Cruz, Rafael MO and Sabourin, Robert and Cavalcanti, George DC and Ren, Tsang Ing},
  journal={Pattern recognition},
  volume={48},
  number={5},
  pages={1925--1935},
  year={2015},
  publisher={Elsevier}
}

@inproceedings{teerapittayanon2016branchynet,
  title={Branchynet: Fast inference via early exiting from deep neural networks},
  author={Teerapittayanon, Surat and McDanel, Bradley and Kung, Hsiang-Tsung},
  booktitle={2016 23rd international conference on pattern recognition (ICPR)},
  pages={2464--2469},
  year={2016},
  organization={IEEE}
}

@article{yang2023dawn,
  title={The dawn of lmms: Preliminary explorations with gpt-4v (ision)},
  author={Yang, Zhengyuan and Li, Linjie and Lin, Kevin and Wang, Jianfeng and Lin, Chung-Ching and Liu, Zicheng and Wang, Lijuan},
  journal={arXiv preprint arXiv:2309.17421},
  year={2023}
}

@article{wei2022chain,
  title={Chain-of-thought prompting elicits reasoning in large language models},
  author={Wei, Jason and Wang, Xuezhi and Schuurmans, Dale and Bosma, Maarten and Xia, Fei and Chi, Ed and Le, Quoc V and Zhou, Denny and others},
  journal={Advances in neural information processing systems},
  volume={35},
  pages={24824--24837},
  year={2022}
}

@article{bhargava2021classification,
  title={Classification and grading of multiple varieties of apple fruit},
  author={Bhargava, Anuja and Bansal, Atul},
  journal={Food Analytical Methods},
  volume={14},
  number={7},
  pages={1359--1368},
  year={2021},
  publisher={Springer}
}

@article{xu2025research,
  title={Research on citrus grading system based on machine vision},
  author={Xu, Miao and Zhang, Xuan and Zhan, ChangJun and Ge, JianYu and Yang, Hua},
  journal={Systems Science \& Control Engineering},
  volume={13},
  number={1},
  pages={2460443},
  year={2025},
  publisher={Taylor \& Francis}
}

@article{astani2022diverse,
  title={A diverse ensemble classifier for tomato disease recognition},
  author={Astani, Mounes and Hasheminejad, Mohammad and Vaghefi, Mahsa},
  journal={Computers and Electronics in Agriculture},
  volume={198},
  pages={107054},
  year={2022},
  publisher={Elsevier}
}

@article{wang2025large,
  title={A large language model for multimodal identification of crop diseases and pests},
  author={Wang, Yiqun and Wang, Fahai and Chen, Wenbai and Lv, Bowen and Liu, Mengchen and Kong, Xiangyuan and Zhao, Chunjiang and Pan, Zhaocen},
  journal={Scientific Reports},
  volume={15},
  number={1},
  pages={21959},
  year={2025},
  publisher={Nature Publishing Group UK London}
}

@inproceedings{szegedy2015going,
  title={Going deeper with convolutions},
  author={Szegedy, Christian and Liu, Wei and Jia, Yangqing and Sermanet, Pierre and Reed, Scott and Anguelov, Dragomir and Erhan, Dumitru and Vanhoucke, Vincent and Rabinovich, Andrew},
  booktitle={Proceedings of the IEEE conference on computer vision and pattern recognition},
  pages={1--9},
  year={2015}
}

@article{simonyan2014very,
  title={Very deep convolutional networks for large-scale image recognition},
  author={Simonyan, Karen and Zisserman, Andrew},
  journal={arXiv preprint arXiv:1409.1556},
  year={2014}
}

@inproceedings{szegedy2016rethinking,
  title={Rethinking the inception architecture for computer vision},
  author={Szegedy, Christian and Vanhoucke, Vincent and Ioffe, Sergey and Shlens, Jon and Wojna, Zbigniew},
  booktitle={Proceedings of the IEEE conference on computer vision and pattern recognition},
  pages={2818--2826},
  year={2016}
}

@inproceedings{chollet2017xception,
  title={Xception: Deep learning with depthwise separable convolutions},
  author={Chollet, Fran{\c{c}}ois},
  booktitle={Proceedings of the IEEE conference on computer vision and pattern recognition},
  pages={1251--1258},
  year={2017}
}

@inproceedings{szegedy2017inception,
  title={Inception-v4, inception-resnet and the impact of residual connections on learning},
  author={Szegedy, Christian and Ioffe, Sergey and Vanhoucke, Vincent and Alemi, Alexander},
  booktitle={Proceedings of the AAAI conference on artificial intelligence},
  volume={31},
  number={1},
  year={2017}
}

@inproceedings{zoph2018learning,
  title={Learning transferable architectures for scalable image recognition},
  author={Zoph, Barret and Vasudevan, Vijay and Shlens, Jonathon and Le, Quoc V},
  booktitle={Proceedings of the IEEE conference on computer vision and pattern recognition},
  pages={8697--8710},
  year={2018}
}

@inproceedings{wu2021cvt,
  title={Cvt: Introducing convolutions to vision transformers},
  author={Wu, Haiping and Xiao, Bin and Codella, Noel and Liu, Mengchen and Dai, Xiyang and Yuan, Lu and Zhang, Lei},
  booktitle={Proceedings of the IEEE/CVF international conference on computer vision},
  pages={22--31},
  year={2021}
}

@article{bao2021beit,
  title={Beit: Bert pre-training of image transformers},
  author={Bao, Hangbo and Dong, Li and Piao, Songhao and Wei, Furu},
  journal={arXiv preprint arXiv:2106.08254},
  year={2021}
}

@inproceedings{messina2022recurrent,
  title={Recurrent vision transformer for solving visual reasoning problems},
  author={Messina, Nicola and Amato, Giuseppe and Carrara, Fabio and Gennaro, Claudio and Falchi, Fabrizio},
  booktitle={International Conference on Image Analysis and Processing},
  pages={50--61},
  year={2022},
  organization={Springer}
}

@article{wang2024qwen2,
  title={Qwen2-vl: Enhancing vision-language model's perception of the world at any resolution},
  author={Wang, Peng and Bai, Shuai and Tan, Sinan and Wang, Shijie and Fan, Zhihao and Bai, Jinze and Chen, Keqin and Liu, Xuejing and Wang, Jialin and Ge, Wenbin and others},
  journal={arXiv preprint arXiv:2409.12191},
  year={2024}
}

@article{garg2023smart,
  title={Smart agriculture: A literature review},
  author={Garg, Disha and Alam, Mansaf},
  journal={Journal of Management Analytics},
  volume={10},
  number={2},
  pages={359--415},
  year={2023},
  publisher={Taylor \& Francis}
}

@article{graves2013generating,
  title={Generating sequences with recurrent neural networks},
  author={Graves, Alex},
  journal={arXiv preprint arXiv:1308.0850},
  year={2013}
}

@article{hinton2015distilling,
  title={Distilling the knowledge in a neural network},
  author={Hinton, Geoffrey and Vinyals, Oriol and Dean, Jeff},
  journal={arXiv preprint arXiv:1503.02531},
  year={2015}
}

@inproceedings{gal2016dropout,
  title={Dropout as a bayesian approximation: Representing model uncertainty in deep learning},
  author={Gal, Yarin and Ghahramani, Zoubin},
  booktitle={international conference on machine learning},
  pages={1050--1059},
  year={2016},
  organization={PMLR}
}

@article{bartlett2008classification,
  title={Classification with a Reject Option using a Hinge Loss.},
  author={Bartlett, Peter L and Wegkamp, Marten H},
  journal={Journal of Machine Learning Research},
  volume={9},
  number={8},
  year={2008}
}

@inproceedings{radford2021learning,
  title={Learning transferable visual models from natural language supervision},
  author={Radford, Alec and Kim, Jong Wook and Hallacy, Chris and Ramesh, Aditya and Goh, Gabriel and Agarwal, Sandhini and Sastry, Girish and Askell, Amanda and Mishkin, Pamela and Clark, Jack and others},
  booktitle={International conference on machine learning},
  pages={8748--8763},
  year={2021},
  organization={PmLR}
}

@inproceedings{lin2017focal,
  title={Focal loss for dense object detection},
  author={Lin, Tsung-Yi and Goyal, Priya and Girshick, Ross and He, Kaiming and Doll{\'a}r, Piotr},
  booktitle={Proceedings of the IEEE international conference on computer vision},
  pages={2980--2988},
  year={2017}
}

@article{yosinski2014transferable,
  title={How transferable are features in deep neural networks?},
  author={Yosinski, Jason and Clune, Jeff and Bengio, Yoshua and Lipson, Hod},
  journal={Advances in neural information processing systems},
  volume={27},
  year={2014}
}

@article{zhang2019lookahead,
  title={Lookahead optimizer: k steps forward, 1 step back},
  author={Zhang, Michael and Lucas, James and Ba, Jimmy and Hinton, Geoffrey E},
  journal={Advances in neural information processing systems},
  volume={32},
  year={2019}
}

@article{krizhevsky2012imagenet,
  title={Imagenet classification with deep convolutional neural networks},
  author={Krizhevsky, Alex and Sutskever, Ilya and Hinton, Geoffrey E},
  journal={Advances in neural information processing systems},
  volume={25},
  year={2012}
}
}


\end{document}